\pdfoutput=1

\documentclass[11pt]{article}

\usepackage[preprint]{ACL2023}

\usepackage{times}
\usepackage{latexsym}

\usepackage[T1]{fontenc}

\usepackage[utf8]{inputenc}

\usepackage{microtype}

\usepackage{inconsolata}

\usepackage{graphicx}

\usepackage{booktabs}
\usepackage{fontawesome}
\usepackage{makecell}   
\usepackage{multirow}
\usepackage[normalem]{ulem}
\usepackage{setspace}
\useunder{\uline}{\ul}{}
\title{A Target-Aware Analysis of Data Augmentation for Hate Speech Detection}

\author{
  Camilla Casula \\
  Fondazione Bruno Kessler \\
  University of Trento \\
  Trento, Italy \\
  \texttt{ccasula@fbk.eu} \\
\And Sara Tonelli \\
  Fondazione Bruno Kessler \\
  Trento, Italy \\
  \\
  \texttt{satonelli@fbk.eu}}
  
\begin{document}
\maketitle
\begin{abstract}
Hate speech is one of the main threats posed by the widespread use of social networks, despite efforts to limit it. Although attention has been devoted to this issue, the lack of datasets and case studies centered around scarcely represented phenomena, such as ableism or ageism, can lead to hate speech detection systems that do not perform well on underrepresented identity groups.

Given the unpreceded capabilities of LLMs in producing high-quality data, we investigate the possibility of augmenting existing data with generative language models, reducing target imbalance. We experiment with augmenting 1,000 posts from the Measuring Hate Speech corpus, an English dataset annotated with target identity information,  adding around 30,000 synthetic examples using both simple data augmentation methods and different types of generative models, comparing autoregressive and sequence-to-sequence approaches. 
We find traditional DA methods to often be preferable to generative models, but the combination of the two tends to lead to the best results. Indeed, for some hate categories such as \emph{origin}, \emph{religion}, and \emph{disability}, hate speech classification using augmented data for training improves by more than 10\% F1 over the no augmentation baseline. This work contributes to the development of systems for hate speech detection that are not only better performing but also fairer and more inclusive towards targets that have been neglected so far.

{\faWarning}\textbf{Warning}: \emph{this paper contains examples of language that may be offensive to some readers.}
\end{abstract}

\section{Introduction}
Online content moderation using automatic systems has been the focus of a large body of research over the past years, due to its importance in tackling the phenomenon of online hate at scale.
Generic hate speech detection models can easily achieve high performance on benchmark datasets, especially for high-resource languages \citep{zampieriSemEval2019TaskIdentifying2019,zampieriSemEval2020Task122020}. Nevertheless, they still present a series of weaknesses. In particular, the creation or maintenance of resources for this task can be problematic due to the relative scarcity of hateful data online \citep{fountaLargeScaleCrowdsourcing2018}, the negative psychological impact the task can have on annotators \citep{riedl2020downsides}, dataset decay or obsolescence, which in turn affects reproducibility \citep{klubickaexamining}, and more.

Aside from their creation process, hate speech datasets can be also problematic in terms of content. For instance,
models have been found to have a tendency to over-rely on specific identity terms, in particular minority group mentions \citep{dixon_measuring_2018, kennedy-etal-2020-contextualizing, rottger-etal-2021-hatecheck, DELAPENASARRACEN2023103433}. 
While this type of problem is especially common in datasets created using keyword sampling \citep{schmidtSurveyHateSpeech2017}, it can still be an issue even for datasets based on different types of data selection \citep{ramponi-tonelli-2022-features}.
Another major issue is related to the representation of minority groups considered as target, which is rather unbalanced, 
potentially affecting the robustness and fairness of hate speech detection systems. 
For example, misogyny and racism have been covered in several datasets \citep{bhattacharya-etal-2020-developing,zeinertAnnotatingOnlineMisogyny2021,guestExpertAnnotatedDataset2021, BOSCO2023103118},
while other phenomena and targets have received much less attention, such as religious hate \citep{ramponi-etal-2022-addressing} or hate against LGBTQIA+ people \citep{chakravarthi2021dataset,locatelli2023cross}, which have only recently started to receive more attention. 
Furthermore, phenomena such as ageism and ableism have been only marginally addressed, and no specific dataset representing these types of offenses has been created. This disparity affects in turn system fairness, because offenses against less-represented targets will typically be classified with a lower accuracy, further impacting communities that are already marginalized \citep{Talat2021DisembodiedML}.

A potential solution that has been proposed for many of the above issues with hate speech detection datasets is the use of synthetic data \citep{vidgenDirectionsAbusiveLanguage2020}. In particular, recent works have experimented with the automatic creation of data using generative large language models, showing it to be a promising solution \citep{juutiLittleGoesLong2020, wullach-etal-2021-fight-fire, dsaExploringConditionalLanguage2021a, hartvigsen-etal-2022-toxigen}, albeit with mixed results \citep{casula-tonelli-2023-generation}. However, no in-depth analysis of the effects of data augmentation for less-represented hate speech targets has been carried out, while it could be beneficial not only to make systems more accurate and robust, but also fairer. Furthermore, synthetic data can in principle mitigate  privacy issues related to collecting and sharing social media data, as well as address the problem of dataset decay, an issue affecting reproducibility \citep{klubickaexamining}: while online messages, especially abusive ones, tend to be deleted over time or can become suddenly inaccessible (see for example Twitter/X change in API policy), synthetic data can be easily generated and reshared. 

Another aspect we want to investigate in this work is a comparison between recent generative language models and more traditional approaches to data augmentation with regards to hate speech detection, which has not been carried out before to our knowledge, although increasing the amount of training data with synthetic examples has been successfully exploited well before the advent of generative large language models \citep{chen-tacl-survey-2023}.

To summarize, we address the following research questions:

\par \textbf{(Q1)} Does data augmentation impact the performance of hate speech detection classifiers differently depending on specific target identities?
\par \textbf{(Q2)} Can information about identity groups in the data augmentation  process help the creation of better and more representative synthetic examples? 
\par \textbf{(Q3)} Do different data augmentation setups and approaches have distinct effects on the performance of models on underrepresented targets?

We answer the above questions through a set of experiments in which we focus on the performance of models by target identity. In addition, we
introduce two novel elements compared to previous work on generative DA: \textit{(i)} we experiment with setups in which we exploit target identity information during generation, attempting to increase the relative representation of scarcely represented targets, and \textit{(ii)} we experiment with instruction-finetuned large language models (LLMs), which have been shown to be able to improve downstream task performances \citep{wei2022finetuned}.
We carry out generation-based data augmentation using 4 different generative models, both with and without access to target identity information. We then perform an evaluation in terms of system performance and fairness. 
We also further investigate potential fairness-related weaknesses of models using the HateCheck test suite \citep{rottger-etal-2021-hatecheck} combined with a manual analysis of generated examples.

The paper is structured as follows: in Section \ref{back} we summarize past works on data augmentation aimed at detecting hate speech. In Section \ref{sec-data} we introduce the Measuring Hate Speech Corpus, which is used in all the experiments. In Section \ref{meth} we detail the methodology adopted in the experiments, in particular how target identify is modeled, the differences between finetuning and few-shot prompting and the prompting layout. In Section \ref{experiment} we describe the experimental setup and the baselines. 
In Section \ref{results} we discuss the experimental results comparing the classification performance of the different setups, while in Section \ref{quality} we perform a qualitative 
analysis by manually annotating generated examples and exploring the weaknesses of our models with the HateCheck test suite \citep{rottger-etal-2021-hatecheck}. We finally discuss the limitations of our study in Section \ref{limit} and summarize our findings and the potential impact of this work in Section \ref{conclusion}.

\section{Background}\label{back}
Although definitions of hate speech in the literature often vary, it is typically defined as a type of offensive speech that specifically targets one or more identity group(s), as opposed to other generically toxic or offensive content \citep{sellars_defining_2016, polettoResourcesBenchmarkCorpora2021}. Since hateful content can have a negative impact on individuals and communities, it is important to develop methods to automatically capture it. 
While over the past years automatic methods for detecting hate speech have achieved very high performances on benchmark datasets (e.g. \citet{zampieriSemEval2019TaskIdentifying2019, zampieriSemEval2020Task122020}), they still exhibit a number of weaknesses \citep{vidgenDirectionsAbusiveLanguage2020}.

The use of synthetic data has been proposed as a way to mitigate some of the known issues with hate speech datasets  \citep{vidgenDirectionsAbusiveLanguage2020}, such as data scarcity \citep{fountaLargeScaleCrowdsourcing2018} and negative psychological impact on annotators \citep{riedl2020downsides}. In particular, data augmentation has emerged as a method to not only potentially mitigate some of the issues with current hate speech and offensive language datasets, but also as a way to improve the performance of hate speech detection systems.

Data augmentation (DA) refers to a family of approaches aimed at increasing the diversity of training data without collecting new samples \citep{feng-etal-2021-survey}. 
While DA is widely used to make models more robust across many machine learning applications \citep{Perez2017TheEO}, it has not been as frequently adopted or researched in NLP \citep{pellicerDataAugmentationTechniques2023, bayerSurveyDataAugmentation2022} until recently, with LLMs being capable of generating realistic text \citep{anaby-tavorNotHaveEnough2020, kumar-etal-2020-data,radfordLanguageModelsAre2019}. 

Indeed, LLMs for data augmentation have been shown to yield performance gains in different NLP tasks such as intent classification \citep{sahu-etal-2022-data}, relation extraction \citep{hu-etal-2023-gda} and  aspect-based sentiment analysis \citep{wang-etal-2023-generative}.  

DA for hate speech detection with generative LLMs has only very recently started to be explored. For example, \citet{juutiLittleGoesLong2020} used GPT-2 \citep{radfordLanguageModelsAre2019} for augmenting toxic language data in extremely low-resource scenarios. Similarly, \citet{wullach-etal-2021-fight-fire} and \citet{dsaExploringConditionalLanguage2021a} augmented toxic language datasets using GPT-2, showing that the addition of large amounts of synthetic data helps classification when starting from datasets containing thousands of labeled instances. 
\citet{fantonHumanintheLoopDataCollection2021} combine GPT-2 and human validation to create counter-narratives covering multiple hate targets. \citet{nouri-2022-data} uses GPT-2 to generate training samples containing offensive text for the task of offensive task detection. 

More recently, \citet{ocampo-etal-2023-depth} have applied data augmentation to increase the number of instances for the minority class in implicit and subtle examples of hate speech, while \citet{casula-tonelli-2023-generation} show that generative data augmentation for hate speech detection  is in some cases challenged by a simple oversampling baseline. \cite{DBLP:conf/emnlp/SarracenRLGP23} propose a variant of vicinal risk minimization \citep{NIPS2000_ba9a56ce} to
generate synthetic samples in the vicinity of the gold examples in a multilingual setting using a multilingual GPT model. Furthermore, \citet{hartvigsen-etal-2022-toxigen} use manually curated (through a human-in-the-loop process) prompts to generate implicitly hateful sequences with GPT-3 \citep{brownLanguageModelsAre2020}.

To our knowledge, no previous study has focused on the problem of augmenting data to address representation issues \textit{for specific targets of hate}. Furthermore, no extensive exploration of the impact of different augmentation strategies and generative models has been presented and evaluated so far on target-based hate speech detection.   

This is also the first work attempting to use instruction-finetuned models for this type of task.

\begin{figure}[t]
    \centering
    \includegraphics[width=0.9\linewidth]{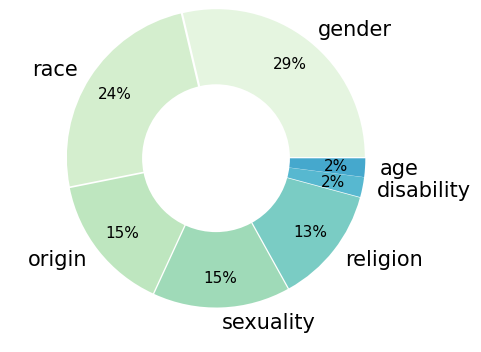}
    \caption{Identity group distribution in the MHS corpus.}
    \label{fig:distr-mhs}
\end{figure}

\begin{table*}[h]
\centering
\resizebox{\textwidth}{!}{%
\begin{tabular}{@{}lll@{}}
\toprule
Text & \multicolumn{1}{l}{Hate speech} & Target identity groups \\ \midrule
I think it was a joke, no black girls eveer dream of such things. & 0 & Gender, Race \\
no one gives a shit f*g & 1 & Sexuality \\ 
People would rather stand up on the bus than sit next to me, hijabi life is awesome & 0 & Religion\\ \bottomrule
\end{tabular}%
}
\label{tab:frasi}
\caption{Examples from the dataset after our aggregation operations, with a hate speech label and a list of target identity groups mentioned or referred to in the text.}
\end{table*}

\section{Data}
\label{sec-data}
For our experiments, we use the Measuring Hate Speech (MHS) Corpus \citep{kennedy_constructing_2020, sachdeva-etal-2022-measuring}, a dataset consisting of social media posts in English from Reddit, Twitter, and YouTube.  

The MHS corpus is annotated according to different levels of hatefulness, from supportive to genocidal speech, offering insight into different aspects of hate speech. In particular, the authors exploit Rasch Measurement Theory to map each example in the dataset to a hate speech score rather than a binary \textit{hateful}/\textit{non-hateful} label. 

However, \citet{kennedy_constructing_2020} also include a binary ``hate speech'' label in the questionnaires to be assigned by annotators. They additionally conduct a comparison between the continuous score and the binary hate speech score, finding that while the continuous measure can better capture the extremity of hate speech, the two are moderately correlated. 
Given the scope of our work, we use the  binary labels instead of the continuous hate speech scores in our experiments, in order to frame the task as classification rather than regression and to be able to test our models on out-of-distribution data (see Section \ref{sec:hatecheck}).

The main characteristic of the MHS dataset that makes it ideal for our study is that it includes labels signaling the presence of  identity groups and sub-groups in texts. Importantly, this annotation is present regardless of hatefulness, resulting in target annotations even for posts containing supportive or counter-speech. 
More specifically, the annotators answered the question `\textit{Is the [comment] directed at or about any individuals or groups based on...'} \citep{sachdeva-etal-2022-measuring}. 

In the freely available version of the MHS dataset\footnote{https://huggingface.co/datasets/ucberkeley-dlab/measuring-hate-speech} we find annotations for seven target identity groups: \textit{race}, \textit{religion}, \textit{origin}, \textit{gender}, \textit{sexuality}, \textit{age}, and \textit{disability}. Their distribution in the data can be seen in Figure \ref{fig:distr-mhs}, which shows how the most widely studied targets of hate speech, \textit{race} and \textit{gender}, are also the most widely represented in the MHS corpus, while some targets such as \textit{age}, \textit{disability}, or \textit{religion} are less frequent.

Given that the MHS dataset, following data perspectivism \citep{Cabitza_Campagner_Basile_2023}, is released with disaggregated annotations, we perform some aggregation operations in order to use it for our experiments, resulting in each example having a unique label and set of targets.
First, we consider each example to be about or targeting all the identity groups identified by at least half of the annotators who annotated it. For example, if out of 5 annotators 3 annotated the target identity group `gender', we will consider this identity group to be the gold target annotation for that example.
Additionally, instead of the hate speech continuous score that is present in the dataset, we use the \texttt{hatespeech} label, which can only take three values (0: \textit{non hatefu}l, 1: \textit{unclear}, 2: \textit{hateful}). We do this in order to frame the task as classification rather than regression for benchmarking purposes, in line with most of the previous work on hate speech detection, in which the task is treated as a classification task. We binarize the three classes by averaging all the annotations for a given post, mapping it to \textit{hateful} if the average score is higher than 1 and to \textit{non hateful} if it is lower.\footnote{While we are aware this does not exploit the most novel and interesting features of the MHS dataset, the exploration of annotator (dis)agreement with regards to data augmentation is beyond the scope of this work, and is left for future research.} 
After this process, we are left with 35,243 annotated posts, of which 9,046 are annotated as containing hate speech.

In the entire corpus, only 48 examples are not associated with any target identity group. Two examples of texts and their annotations from the processed corpus we use are shown in Table 1.\footnote{The original messages from the dataset were changed slightly to avoid using user content in this paper. Additionally, slurs are manually obfuscated by us, following the guidelines by  \cite{nozza-hovy-2023-state}.}

\section{Methodology}\label{meth}

\begin{figure*}[h]
     \centering
     \includegraphics[width=0.75\linewidth]{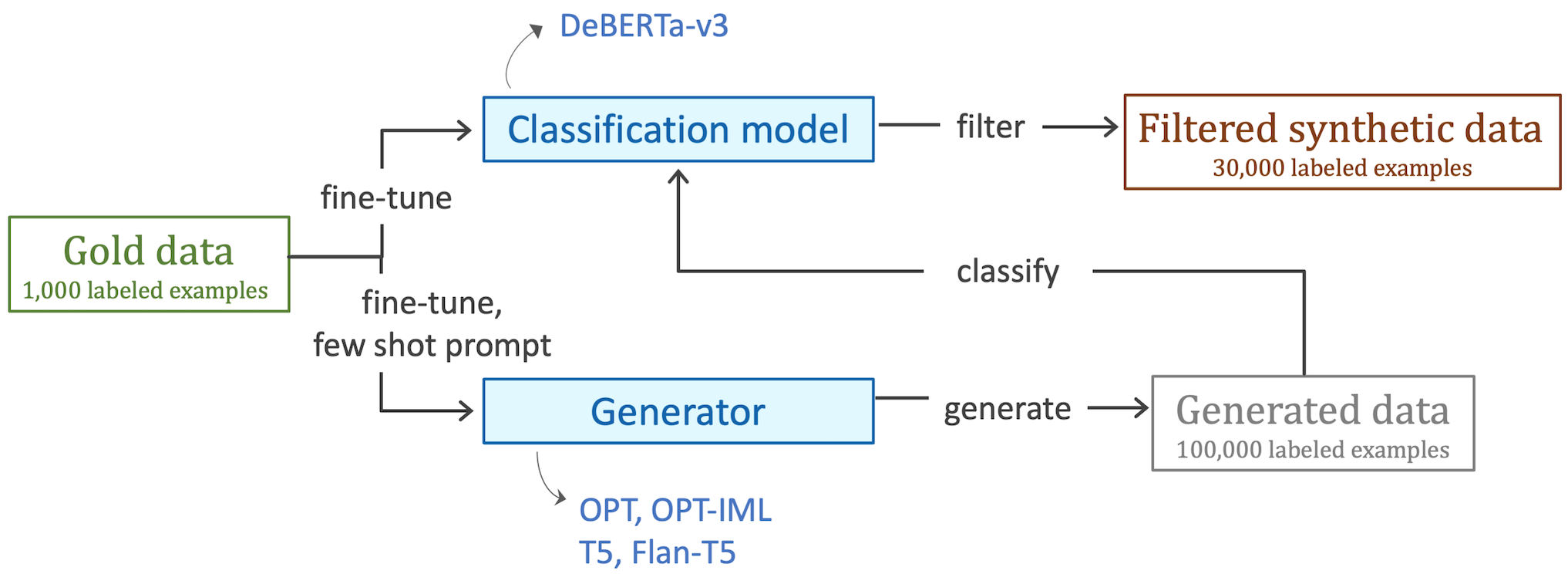}
     \caption{The generative DA pipeline, from gold data to filtered syntetic data.}
     \label{fig:da-pipeline}
 \end{figure*}

In our experiments, we implement a data augmentation pipeline inspired by \cite{anaby-tavorNotHaveEnough2020}, which has also been used in other work \citep{wullach-etal-2021-fight-fire, casula-tonelli-2023-generation}. The pipeline is displayed in Figure \ref{fig:da-pipeline}. Starting from a small set of \textit{Gold data} from the MHS corpus, a \textit{Generator} is employed to augment them by generating synthetic examples with the corresponding label (hateful or not) using either \textit{finetuning} or \textit{few-shot prompting} (see Section \ref{finetune}). Since the labels associated with the \textit{Generated data} may not be accurate, given that generative models cannot always preserve the desired labels \citep{kumar-etal-2020-data}, a subsequent filtering step is used in order to maximize the chances of label correctness. In order to create a model for filtering generated texts, the same gold data is used also to fine-tune a binary classifier that assigns a hateful/non hateful label to the generated data. We keep only the synthetic examples where there is a match between the label assigned during generation and by the \textit{Classification model} to filter out examples that are likely to have an inconsistent labels. 
The \textit{Filtered synthetic data} is then used to train a hate speech classifier that we evaluate for the task performance in general and then on specific hate targets.  

We detail below the variants we test for each step of the pipeline.

\subsection{Generative Models}

For the \textbf{Generator} step, we experiment with four different transformer models: OPT \citep{zhang2022opt} and T5 \citep{raffel2020t5} and their instruction-finetuned counterparts: OPT-IML \citep{iyer2022opt} and Flan-T5 \citep{chung2022flant5}. We choose to only use openly available models for our experiments to favor reproducibility. The selected models allow us also to compare a decoder-only model (OPT) with an encoder-decoder model (T5), which to our knowledge has not been done in previous studies on this type of data augmentation \citep{casula-tonelli-2023-generation,azam-etal-2022-exploring}. 

Another aspect we want to investigate is the  performance of instruction-finetuned models compared to their standard version, since recent works showed that  instruction-tuning can improve generalization to unseen tasks \citep{chung2022flant5}. For this reason, we include OPT-IML and Flan-T5 beside OPT and T5.

We use the 1.3B parameter version of OPT and OPT-IML and the Large version of T5 and Flan-T5 (770M).\footnote{While this means there is a disparity between the decoder-only and the encoder-decoder model sizes we use, 1.3B is the smallest available model size for OPT-IML, and finetuning Flan-T5 3B, the next model size available for the encoder-decoder architecture, was beyond our computing capacity.}

\subsection{Finetuning vs Few-Shot Prompting with and without Target Group Identity Information}\label{finetune}
A large number of works on data augmentation based on generative models rely on finetuning a model on a small set of gold data, and then generating new data with the finetuned model, encoding the label information within the text sequences in some form (e.g. \citet{anaby-tavorNotHaveEnough2020, kumar-etal-2020-data}). Other works use few-shot demonstration-based prompting, in which the pre-trained model is prompted with one or more sequences similar to what the model is expected to generate, with no finetuning (e.g. \citet{hartvigsen-etal-2022-toxigen, azam-etal-2022-exploring, ashida-komachi-2022-towards}).
We experiment with both strategies for each transformer model. 

Since our research questions revolve around the impact that hate target information can have on data augmentation, finetuning and few-shot demonstration-based prompting are further tested in two variants: with and without mentioning the \textbf{target identity information}. Our hypothesis is that the inclusion of this kind of information might help in generating more varied data with regards to identity group mentions for both hateful and non-hateful messages. By generating target-specific examples also for the non-hateful class, we ideally aim at implicitly contrasting identity term bias. In order to do this, we encode target identity information into the prompts given to the models in various ways. The values used for ‘target’ are the identity group names in the MHS dataset, reported in Section \ref{sec-data}.

\subsubsection{Finetuning (FT)} 

For finetuning, we follow an approach similar to that of \citet{anaby-tavorNotHaveEnough2020}, in which a generative LLM is fine-tuned on annotated sequences that are concatenated with labels. At generation time, the desired label information is fed into the model, and the model is expected to generate a sequence belonging to the specified class. 
We discuss the details of the formatting of the label information in Section \ref{subsubsec:prompting}.

This method has the upside of theoretically being more likely to generate examples that are closer to the original distribution of the data to be augmented. However, this can also be a downside, if the desired effect is increasing the variety of the data. In addition, finetuning is more computationally expensive than few-shot prompting.
For models fine-tuned with target identity information, given that each sequence can be associated with more than one target (in cases of intersectional hate speech for instance), the label-encoding sequence will include all targets in that post.

\subsubsection{Few-shot prompting (FS)}
Following the large amount of works focusing on few-shot demonstration-based instructions, especially with instruction-finetuned models \citep{iyer2022opt, chung2022flant5}, we also experiment with demonstration-based prompting, in which the models are shown 3 examples belonging to the desired label (and target identity, if available), and then asked to produce a new one. 

With models exploiting target identity information for few-shot prompting, we associate the desired label and target with 3 sequences. For instance, if the model is expected to generate a non-hateful post about gender, we select 3 sequences that are annotated in the gold data as non-hateful and about gender.

\subsubsection{Formatting}
\label{subsubsec:prompting}

As regards formatting, we aim at using the same type of prompting layout across experiments. We choose to use prompting sequences in natural language, given that they have been found to lead to generally more realistic generated examples \citep{casula-tonelli-2023-generation}. In order to find prompts in natural language that could be leveraged by our models, we consulted the FLAN corpus  \citep{wei2022finetuned}, which is part of the finetuning data of both FLAN-T5 and OPT-IML. Among the instruction templates, we find one of the CommonGen templates \citep{lin-etal-2020-commongen} to fit with our aims: `\textit{Write a sentence about the following things: [concepts], [target]}'. We reformulate it to obtain a prompting sequence that reflects our application, and can be exploited by instruction-finetuned models:
         `Write a [$\emptyset$/ \textit{hateful}] social media post [$\emptyset$/ about \textit{t}]',
where $t$ is a target identity.
Table \ref{tab:ex-prompt} presents the sequences and prompts used for training and prompting our models. 
\begin{table}[h!]
\centering
\resizebox{\linewidth}{!}{%
\begin{tabular}{ll}
\toprule
 Target & Write a [$\emptyset$ | hateful] social media post about \{target\}: \{text\} \\
\midrule
 No target & Write a [$\emptyset$ | hateful] social media post: \{text\} \\
 \bottomrule

\end{tabular}%
}

 \caption{Templates used for fine-tuning and prompting generative models during the generation step.}\label{tab:ex-prompt}

\end{table}

To summarize, we use four Transformer models (OPT, OPT-IML, T5 and Flan-T5) to generate synthetic data using finetuning or few-shot prompting, either with access to target information or not, for a total of sixteen different strategies that we evaluate in our experiments.

\section{Experimental Setup}\label{experiment}
For all experiments, following previous work on data augmentation \citep{anaby-tavorNotHaveEnough2020, casula-tonelli-2023-generation}, we simulate a setup in which we have a small amount of gold data available prior to augmentation (see Figure \ref{fig:da-pipeline}). We randomly select 1,000 gold examples from the MHS corpus, as we deem it a realistically small dataset size for a hate speech detection corpus, after looking at the hate speech dataset review by \citet{vidgenDirectionsAbusiveLanguage2020}.
Our goal is to create a larger dataset out of the starting 1,000 examples. 
Given that the `natural' size of the Measuring HS dataset is 35k examples, we aim for 30k new annotated examples (equally split between the labels) to use in augmentation, which will result in a 31k example dataset for each setup.

\citet{wullach-etal-2021-fight-fire}, using a similar DA approach, preserve around 1/3 of the generated examples after filtering. 
We therefore generate around 3 times as many examples as we need, setting the total number of generated examples for each setup to 100,000, equally divided into hateful and not hateful. For each setup, we thus prompt models to generate 50k examples for each class. 

Given that our focus is on different targets of hate, we aim at investigating the impact of their representation in the data on model performance. Specifically, since their distribution in the MHS corpus is highly imbalanced, as seen in Section \ref{sec-data}, we hypothesize that their representation might influence the performance of models. Because of this, we choose to equally augment each target identity category (gender, race, origin, sexuality, religion, disability, and age). Indeed, for models that rely on target identity information, out of the 50k generated instances for each class, we generate 1/7 for each target identity category (7,140 newly generated sequences for each target identity group).

Once we generate 100,000 examples, we filter them by predicting their hate speech label using a DeBERTa-v3 Large classifier \citep{he2023debertav3}

 finetuned on the initial 1,000 gold examples.\footnote{We replicate all the experiments also with RoBERTa Large \citep{liu_roberta_2019}. The results are in line with those obtained using DeBERTa, therefore they are not reported in this paper.} We only preserve the examples for which the classifier label assignment matches the desired label that was in the model input at generation time, in line with previous work that used this kind of filtering for synthetic sequences.
If less than 15k generated sequences for a given label pass filtering, we preserve the examples that did pass filtering, and proceed with the rest of the pipeline. In some setups, this means that we end up with fewer synthetic examples than initially expected. We discuss this more in depth in Section \ref{results}.

We test the quality of the synthetic data extrinsically, by using it in addition to the initial available gold data for training classifiers aimed at detecting the presence of hate speech for specific targets. The hypothesis is that better-quality synthetic data will lead to better performance of models trained on data augmented with it, when compared with the performance of models trained on gold data alone.

\subsection{Implementation details with Generative DA}

For all of our experiments, we employ the HuggingFace library \citep{wolf-etal-2020-transformers}. All the hyperparameters we use that are not specified in this section are the default ones from their \texttt{TrainingArguments} class. 
The DeBERTa classifiers we use as baselines and for filtering are trained on 5 epochs.

We use $LR=5e-6$ and a batch size of 16 for training.
We fine-tune T5 Flan-T5, OPT and OPT-IML with batch 16 and $LR=1e-3$. For generation, we set each iteration to generate 10 sequences for both finetuned and demonstration-based prompting, using \textit{top-p}=0.9 and setting min and max lengths of generated sequences to 5 and 150 tokens, respectively.
All the classifiers that are trained on augmented data are trained for 3 epochs (given that they are trained on more data, they require less epochs to converge), again with batch size 16 and LR=$5e-6$. In this case, at the end of training, we preserve the model from the epoch with the lowest evaluation cross-entropy loss.

The random seeds we used for shuffling, subsampling the gold data, and initializing both generative and classification models are 522, 97, 709, 16, and 42. These were chosen randomly.
Finetuning of all classifiers and generative models, including baselines and models trained on augmented data, took 50 hours, of which 45 on a Nvidia V100 GPU and 5 on a Nvidia A40. Inference time for generating all of the sequences (a total of 8 million generated texts) took $\sim$500 hours total.

\subsection{System comparison}

We compare all of our models with \textbf{Easy Data Augmentation (EDA)} \citep{wei-zou-2019-eda} in its implementation by \citet{marivate2020improving}. EDA consists of four operations: synonym replacement using WordNet \citep{miller-1992-wordnet}, random insertion, random swap, and random deletion of tokens. Similarly to our other setups, we produce 30k new sequences with EDA, of which 7,500 with each operation, on the initial 1,000 examples in each fold. We then also experiment with the mixture of EDA and generative DA, in which instead of augmenting the initial gold data with 30k synthetic sequences obtained with EDA or generative DA, we randomly select 15k examples of LLM-generated texts and 15k examples of EDA-perturbed examples and concatenate them. 

We also implement \textbf{two baselines} using DeBERTa: \textit{i)} the classifier finetuned on the starting 1k gold examples, and \textit{ii)} the same classifier finetuned on an oversampled version of the training data (repeating the initial 1k sequences until we get to 31k, the size of the augmented setups), which has been found effective even in cross-dataset scenarios with regards to data augmentation for offensive language data \citep{casula-tonelli-2023-generation}.

\begin{table*}[ht]
\centering
\resizebox{0.9\linewidth}{!}{%
\addtolength{\tabcolsep}{-0.2em}
\begin{tabular}{@{}ccc|ll|lllllll|l@{}}
\toprule
                        & & & \multirow{2}{*}{M-F$_1$}& \multirow{2}{*}{Hate-F$_1$}& \multicolumn{7}{c|}{Hate-F$_1$}&  \multirow{2}{*}{n(h)}\\
                         &                       &     &            &           & Gender& Race& Origin& Sexuality& Religion& Disability& Age&  \\

\midrule
\multicolumn{3}{l|}{\textbf{No augmentation}}  & .773$^{.02}$            & .652$^{.03}$            & .635$^{.02}$            & .696$^{.04}$            & .497$^{.05}$            & .756$^{.03}$            & .485$^{.12}$            & .698$^{.03}$            & .545$^{.04}$         &      \\
\multicolumn{3}{l|}{\textbf{Oversampling}}  & .773$^{.02}$            & .653$^{.04}$            & .652$^{.05}$            & .740$^{.02}$$\star$          & .568$^{.05}$$\star$         & .787$^{.02}$$\star$         & .571$^{.03}$$\star$         & .732$^{.04}\diamond$           & .555$^{.06}$         &      \\
\multicolumn{3}{l|}{\textbf{EDA}} & \textbf {.799$^{.01}$$\star$} & \textbf {.714$^{.01}$$\star$} & \textbf {.687$^{.01}$$\star$} & \textbf {.771$^{.02}$$\star$} & \textbf {.582$^{.03}$$\star$} & \textbf {.806$^{.01}$$\star$} & \textbf {.601$^{.02}$$\star$} & \textbf {.799$^{.02}$$\star$} & .589$^{.06}$$\star$      & 15k  \\
\midrule
Model                    &                 & Target &                        &                        &                        &                        &                        &                        &                        &                        &                     &      \\
\midrule
\multirow{4}{*}{OPT}     & \multirow{2}{*}{FT}   & Y & .783$^{.00}$$\star$             & .683$^{.01}$$\star$         & .653$^{.03}\diamond$           & .740$^{.02}$$\star$          & .556$^{.05}$$\star$         & .779$^{.02}\diamond$           & .535$^{.07}$            & .777$^{.02}$$\star$         & .587$^{.06}\diamond$        & 0.5k \\
                         &                       & N  & .774$^{.04}$            & .652$^{.07}$            & .634$^{.05}$            & .707$^{.06}$            & .505$^{.06}$            & .738$^{.10}$             & .461$^{.07}$            & .690$^{.11}$             & .590$^{.10}\diamond$          & 15k  \\
                         & \multirow{2}{*}{FS} & Y & .782$^{.01}\diamond$           & .691$^{.02}$$\star$         & .667$^{.02}$$\star$         & .750$^{.02}$$\star$          & .553$^{.04}$$\star$         & .790$^{.01}$$\star$          & .546$^{.05}\diamond$           & .791$^{.02}\diamond$           & .582$^{.07}\diamond$        & 11k  \\
                         &                       & N  & .791$^{.01}$$\star$         & .700$^{.01}$$\star$           & .675$^{.02}$$\star$         & .758$^{.02}$$\star$         & .561$^{.02}$$\star$         & .791$^{.02}$$\star$         & .555$^{.07}\diamond$           & .776$^{.03}$$\star$         & .597$^{.05}$$\star$      & 15k  \\
\midrule
\multirow{4}{*}{\begin{tabular}[c]{@{}c@{}}OPT\\IML\end{tabular}} & \multirow{2}{*}{FT}   & Y & .789$^{.01}$$\star$         & .681$^{.02}$$\star$         & .661$^{.02}$$\star$         & .720$^{.05}$             & .516$^{.09}$            & .789$^{.01}$$\star$         & .493$^{.05}$            & .735$^{.04}\diamond$           & .579$^{.06}\diamond$        & 15k  \\
                         &                       & N  & .796$^{.01}$$\star$         & .690$^{.02}$$\star$          & .674$^{.03}$$\star$         & .738$^{.02}$$\star$         & .500$^{.07}$              & .791$^{.02}$$\star$         & .488$^{.10}$             & .723$^{.09}$            & .593$^{.10}\diamond$         & 15k  \\
                         & \multirow{2}{*}{FS} & Y & .789$^{.01}\diamond$           & .698$^{.01}$$\star$         & .672$^{.02}$$\star$         & .757$^{.02}$$\star$         & .563$^{.03}$$\star$         & .798$^{.02}$$\star$         & .552$^{.07}\diamond$           & .780$^{.03}$$\star$          & .577$^{.07}$         & 11k  \\
                         &                       & N  & .792$^{.01}$$\star$         & .699$^{.01}$$\star$         & .673$^{.02}$$\star$         & .755$^{.02}$$\star$         & .564$^{.03}$$\star$         & .795$^{.01}$$\star$         & .558$^{.06}\diamond$           & .772$^{.04}$$\star$         & \textbf{.604}$^{.05}$$\star$      & 15k  \\
\midrule
\multirow{4}{*}{T5}      & \multirow{2}{*}{FT}   & Y & .792$^{.01}$$\star$         & .696$^{.02}$$\star$         & .667$^{.02}$$\star$         & .753$^{.02}$$\star$         & .567$^{.04}$$\star$         & .795$^{.02}$$\star$         & .566$^{.05}\diamond$           & .771$^{.03}$$\star$         & .584$^{.09}$         & 12k  \\
                         &                       & N  & .789$^{.01}$$\star$         & .684$^{.01}$$\star$         & .660$^{.02}$$\star$          & .731$^{.03}\diamond$           & .536$^{.02}$$\star$         & .784$^{.01}$$\star$         & .523$^{.08}$            & .748$^{.04}$$\star$         & .592$^{.07}\diamond$        & 10k  \\
                         & \multirow{2}{*}{FS} & Y & .786$^{.01}\diamond$           & .682$^{.02}$$\star$         & .674$^{.03}$$\star$         & .738$^{.02}$$\star$         & .500$^{.07}$              & .791$^{.02}$$\star$         & .488$^{.10}$             & .723$^{.09}$            & .593$^{.10}$$\star$       & 11k  \\
                         &                       & N  & .798$^{.01}$$\star$         & .700$^{.02}$$\star$         & .666$^{.02}$$\star$         & .756$^{.02}$$\star$         & .559$^{.07}\diamond$           & .793$^{.01}$$\star$         & .573$^{.05}\diamond$           & .774$^{.03}$$\star$         & .596$^{.04}$$\star$      & 15k  \\
\midrule
\multirow{4}{*}{\begin{tabular}[l]{@{}c@{}}FLAN\\T5\end{tabular}} & \multirow{2}{*}{FT}   & Y & .792$^{.01}$$\star$         & .696$^{.01}$$\star$         & .669$^{.01}$$\star$         & .752$^{.01}$$\star$         & .559$^{.03}$$\star$         & .792$^{.02}$$\star$         & .574$^{.05}\diamond$           & .767$^{.03}$$\star$         & .600$^{.07}\diamond$ & 14k  \\
                         &                       & N  & .793$^{.01}$$\star$         & .691$^{.01}$$\star$         & .672$^{.02}$$\star$         & .737$^{.03}$$\star$         & .544$^{.05}$            & .790$^{.01}$$\star$          & .520$^{.08}$            & .750$^{.04}$$\star$          & .597$^{.08}$         & 10k  \\
                         & \multirow{2}{*}{FS} & Y & .786$^{.00}$$\star$             & .684$^{.01}$$\star$         & .651$^{.02}$            & .743$^{.02}$$\star$         & .558$^{.04}$$\star$         & .778$^{.01}$$\star$         & .536$^{.04}$            & .744$^{.04}$$\star$         & .590$^{.10}$           & 0.3k \\
                         &                       & N  & .774$^{.02}$            & .662$^{.04}$            & .637$^{.04}$            & .709$^{.06}$            & .509$^{.09}$            & .765$^{.03}$            & .490$^{.09}$             & .724$^{.06}$            & .583$^{.09}$         & 0.3k \\ 
\bottomrule
\end{tabular}
}
\caption{DeBERTa results (macro-F$_1$ and hate-class F$_1$) with generative DA, averaged over 5 runs $^{\pm stdev}$, overall and by target (\textit{Ge}nder, \textit{Ra}ce, \textit{Or}igin, \textit{Se}xuality, \textit{Re}ligion, \textit{Di}sability, and \textit{Ag}e). 
Statistical significance is calculated against the \textit{no augmentation} baseline. $\star$: highly statistically significant ($\tau=0.2$), $\diamond$: statistically significant ($\tau=0.5$). \textit{n(h)} = number of \textit{hateful} synthetic examples preserved after filtering.}
\label{tab:deberta-results-gen}
\end{table*}

\section{Results and Discussion}\label{results}
In this section we report the results of our experiments averaged across 5 data folds. We test statistical significance using Almost Stochastic Order (ASO) \citep{dror-etal-2019-deep, del2018optimal}, as implemented by \citet{ulmer2022deep}.

\subsection{Generative DA}

We report in Table \ref{tab:deberta-results-gen} the results of our experiments using generative DA and compared with EDA and the two baselines described above. The classification performance is evaluated globally in terms of macro-F$_1$ and minority (hate) class F$_1$ and for each target identity category as hate class F$_1$, so that the impact of synthetic data can be examined on a per-target basis.

Even from the \textit{no augmentation} baseline, it is clear that performance can vary greatly across targets, with up to 27\% hate-F$_1$ differences between them. In particular, the model appears to struggle with posts about \textit{origin}, \textit{religion}, and \textit{age}, while, although underrepresented, posts about \textit{disability} tend to be classified more accurately. This suggests that performance might also be influenced by factors other than the representation of targets in the dataset, such as how broad a target category is. For instance, \textit{origin} can include any type of discrimination based on geographical origin, including specific countries, and \textit{religion} can encompass any type of religious discourse, although each religion is often targeted through specific offense types \citep{ramponi-etal-2022-addressing}. This makes classification challenging, especially for systems mostly relying on lexicon. This shows also how relevant it is to assess performance on targets separately, as examples referring to different target identity groups might pose different challenges for classification.

Most of the models trained on generation-augmented data outperform the \textit{no augmentation} baseline across targets, with different improvements based on target identity group (\textit{origin}, \textit{religion}, and \textit{age} in particular). Strikingly, however, EDA performs better than all generation-based DA configurations, regardless of prompting type or access to target information, for all targets but \textit{age}. 
While performance gains are similar between EDA and the best generation-based setup compared to the baseline (+.026 and +.025 M-F$_{1}$ respectively),
EDA appears to lead to slightly better performance in terms of minority class F$_1$ (+.062 against +.048), at a small fraction of the computational cost of the generation-based approaches. 
This is reflected on the performance per target identity, in which EDA outperforms generative DA across all but one target, \textit{age}, the least represented one in the data.
We hypothesize EDA is effective because small perturbations can make models more robust, especially with regards to the \textit{hateful} class, while generative models might increase performance, but they might also be more likely to inject noise.

The impact of finetuning vs. few-shot prompting seems model-dependent, with differences across models also regarding the impact of target information. For all but OPT-IML, finetuning approaches tend to favor the inclusion of target information, albeit with relatively minor differences. Interestingly, the amount of synthetic examples labeled as \textit{hateful} (reported in Table \ref{tab:deberta-results-gen} as \textit{n(h)}) that pass filtering does not appear to strongly impact the performance of models trained on synthetic data, indicating that potentially even just a few hundred synthetic examples can positively impact generalization.
This could also indicate that even just the addition of non-hateful synthetic examples can help models to generalize.

\begin{table*}[h]
\centering
\resizebox{0.9\linewidth}{!}{%
\addtolength{\tabcolsep}{-0.2em}
\begin{tabular}{@{}ccc|ll|lllllll@{}}
\toprule
                        & & & \multirow{2}{*}{M-F$_1$}& \multirow{2}{*}{Hate-F$_1$}& \multicolumn{7}{c}{Hate-F$_1$}\\
                         &                       &     &            &           & Gender& Race& Origin& Sexuality& Religion& Disability& Age\\
 \midrule
 \multicolumn{3}{l|}{\textbf{No augmentation}}  & .773$^{.02}$            & .652$^{.03}$            & .635$^{.02}$            & .696$^{.04}$            & .497$^{.05}$            & .756$^{.03}$            & .485$^{.12}$            & .698$^{.03}$            & .545$^{.04}$         \\
\multicolumn{3}{l|}{\textbf{EDA}} & .799$^{.01}$ & .714$^{.01}$ & .687$^{.01}$ & .771$^{.02}$ & .582$^{.03}$ & .806$^{.01}$ & .601$^{.02}$ & .799$^{.02}$ & .589$^{.06}$ \\
\midrule
Model &  & Tar &  &  &  &  &  &  &  &  &  \\
\midrule
\multirow{4}{*}{\begin{tabular}[c]{@{}c@{}}OPT \\ +\\ EDA\end{tabular}} & \multirow{2}{*}{FT} & Y & .777$^{.02}$ & .698$^{.02}$ & .679$^{.02}$ & .759$^{.01}$ & .567$^{.05}$ & .795$^{.01}$ & .593$^{.03}$ & .801$^{.03}$ & .578$^{.07}$ \\
 &  & N & .792$^{.02}$ & .711$^{.02}$ & .687$^{.02}$ & .768$^{.02}$ & .586$^{.06}$ & .806$^{.01}$ & .599$^{.03}$ & \textbf{.812}$^{.02}\diamond$ & .588$^{.06}$ \\
 & \multirow{2}{*}{FS} & Y & .788$^{.02}$ & .707$^{.01}$ & .684$^{.01}$ & .767$^{.01}$ & .579$^{.04}$ & .803$^{.01}$ & .588$^{.03}$ & .803$^{.02}$ & .588$^{.06}$ \\
 &  & N & .795$^{.03}$ & .715$^{.03}$ & .689$^{.03}$ & .774$^{.02}$ & .595$^{.04}$ & .808$^{.03}$ & .617$^{.03}\diamond$ & .802$^{.04}$ & \textbf{.631$^{.07}\diamond$} \\
 \midrule
\multirow{4}{*}{\begin{tabular}[c]{@{}c@{}}OPT-IML\\ +\\ EDA\end{tabular}} & \multirow{2}{*}{FT} & Y & .788$^{.01}$ & .709$^{.01}$ & .687$^{.02}$ & .767$^{.01}$ & .577$^{.04}$ & .808$^{.02}$ & .611$^{.01}\diamond$ & .797$^{.03}$ & .584$^{.04}$ \\
 &  & N & .791$^{.01}$ & .710$^{.01}$ & .687$^{.01}$ & .766$^{.01}$ & .580$^{.05}$ & .805$^{.02}$ & .597$^{.03}$ & .800$^{.01}$ & .597$^{.05}$ \\
 & \multirow{2}{*}{FS} & Y & .789$^{.02}$ & .709$^{.02}$ & .685$^{.02}$ & .769$^{.01}$ & .584$^{.05}$ & .805$^{.02}$ & .615$^{.03}\diamond$ & .795$^{.04}$ & .597$^{.06}$ \\
 &  & N & .791$^{.01}$ & .711$^{.01}$ & .689$^{.02}$ & .766$^{.01}$ & .588$^{.04}$ & .809$^{.01}$ & .601$^{.02}$ & .807$^{.02}$ & .596$^{.06}$ \\
 \midrule
\multirow{4}{*}{\begin{tabular}[c]{@{}c@{}}T5\\  +\\ EDA\end{tabular}} & \multirow{2}{*}{FT} & Y & \textbf{.805$^{.01}$} & \textbf{.722$^{.01}\diamond$} & .695$^{.01}\diamond$ & \textbf{.778$^{.01}$} & \textbf{.596}$^{.03}$ & .815$^{.01}\diamond$ & \textbf{.628$^{.03}\diamond$} & .808$^{.03}$ & .588$^{.08}$ \\
 &  & N & .799$^{.00}$& .716$^{.01}$ & \textbf{.696$^{.01}\diamond$} & .772$^{.01}$ & .589$^{.03}$ & .810$^{.02}$ & .610$^{.03}$ & .800$^{.02}$ & .617$^{.07}$ \\
 & \multirow{2}{*}{FS} & Y & .796$^{.01}$ & .715$^{.01}$ & .689$^{.02}$ & .776$^{.01}$ & .582$^{.04}$ & .807$^{.02}$ & .611$^{.03}$ & .809$^{.02}\diamond$ & .618$^{.06}$ \\
 &  & N & .793$^{.01}$ & .712$^{.01}$ & .691$^{.02}$ & .771$^{.01}$ & .586$^{.04}$ & .809$^{.01}$ & .602$^{.01}$ & .803$^{.02}$ & .619$^{.03}\diamond$ \\
 \midrule
\multirow{4}{*}{\begin{tabular}[c]{@{}c@{}}Flan-T5\\ +\\ EDA\end{tabular}} & \multirow{2}{*}{FT} & Y & .803$^{.01}$ & .718$^{.01}$ & .690$^{.01}$ & .774$^{.02}$ & .586$^{.04}$ & .813$^{.01}\diamond$ & .609$^{.04}$ & .801$^{.03}$ & .597$^{.07}$ \\
 &  & N & .794$^{.01}$ & .712$^{.01}$ & .690$^{.02}$ & .766$^{.01}$ & .585$^{.04}$ & .811$^{.01}$ & .600$^{.02}$ & .794$^{.03}$ & .582$^{.06}$ \\
 & \multirow{2}{*}{FS} & Y & .788$^{.02}$ & .707$^{.02}$ & .685$^{.03}$ & .767$^{.02}$ & .574$^{.05}$ & .807$^{.01}$ & .598$^{.03}$ & .797$^{.03}$ & .590$^{.06}$ \\
 &  & N & .799$^{.02}$ & .718$^{.02}$ & .691$^{.03}$ & .777$^{.02}$ & .582$^{.03}$ & \textbf{.816}$^{.02}\diamond$ & .619$^{.03}\diamond$ & .810$^{.03}$ & .615$^{.08}$\\
 \bottomrule
\end{tabular}
}
\caption{DeBERTa results of generative DA + EDA overall and by target, averaged over 5 runs $^{\pm stdev}$. Statistical significance is calculated against the results obtained with EDA. $\diamond$: statistically significant against EDA alone ($\tau=0.5$).}
\label{tab:deberta-results-gen+eda}
\end{table*}

\subsection{Mixture of Generative DA and EDA}
\label{sec:mixture}
Since models trained on EDA-augmented data outperform models trained only on generation-augmented sequences, we also experiment with the mixture of the two methods, with 15k synthetic examples created using each of them. In Table \ref{tab:deberta-results-gen+eda} we report the results of these experiments.

Overall, it appears that the combination of EDA and generative DA can outperform each of the two methods separately, with some differences across models, augmentation setups, and target groups. The setup with EDA and the T5 model finetuned with target information leads to statistically significant hate-F$_1$ gains over EDA both overall and on the \textit{gender}, \textit{sexuality}, and \textit{religion} targets. In addition, the classification of \textit{origin}, \textit{religion}, and \textit{disability} improves by around or over 10\% M-F$_1$ over the \textit{no augmentation} baseline, showing the potential of this DA setup. However, the impact of finetuning vs. few-shot prompting still appears to be model-dependent, similarly to the impact of target identity information in the prompts.

In general, the high computational cost of generative approaches might not always justify their use against simpler yet effective DA approaches such as EDA in low-resource scenarios. 
Nevertheless, the combination of the two methods can outperform each method alone, and we hypothesize that it may be due to the fact that the gains are complementary: while EDA can make models robust to small perturbations such as word order changes, generative DA could be better at increasing lexical variety.

\section{Qualitative Analysis}\label{quality}
In this section, we look into  the synthetically generated texts and the models trained on them from a qualitative point of view. First we carry out a manual annotation on the generated texts to compare the different settings in terms of realism, target identity group assignment correctness and label consistency. Our goal is to assess whether these three dimensions in the generated data correlate with classifier performance. Then, we turn to the HateCheck test suite \citep{rottger-etal-2021-hatecheck}, a hate-speech specific test set which includes examples divided by targets aimed at exploring the weaknesses of hate speech models, especially their out-of-distribution generalisation. 
Indeed, HateCheck targets do not exactly overlap with the target categories of the MHS dataset, thus providing a complementary view on our models' performance.

\subsection{Manual Annotation}
A total of 1,120 generated texts filtered with DeBERTa were annotated by two  annotators with a background in linguistics and experience in hate speech research. For each combination of finetuning/prompting/target presence (16 setups), they annotated 70 examples, evenly distributed across labels and, in the experiments that used them, targets. 

In particular, for each setting for generative DA \textit{with target information}, annotators were asked to annotate synthetic examples by specifying the following dimensions:
\begin{itemize}
    \item Label: whether the content of the text is hateful or not,
    \item Target Match: whether the target mentioned in the text matches with the target identity category given in input to the generative model,
    \item Realism: whether the message appears realistic and could have been plausibly written by a human.
\end{itemize}

For the examples generated \textit{without access to target information}, the Target dimension was not annotated.

Consider for example the following sentence, generated giving `age' as target information:

\vspace{0.4cm}
\textit{`F*ckin white men are trashy like a muthaf*cker'.}
\vspace{0.2cm}

In this case, Label would be `\textit{hateful}', Realism would be `\textit{Yes}' but Target would be `\textit{No}', because the target identity category of the generated example is `race' and not `age'.
Inter-annotator agreement was calculated using Krippendorff's alpha on 10\% of the manually analyzed data. The annotators showed moderate agreement with regards to label correctness ($\alpha$ = 0.76), while the scores were higher for target identity group matches ($\alpha$ = 0.83) and realism ($\alpha$ = 0.82).

\begin{table}[h]
\centering
\resizebox{\linewidth}{!}{%
\addtolength{\tabcolsep}{-0.2em}
\begin{tabular}{@{}ccc|ccc|cc@{}}
\toprule
Model &  & Target & Label & Tar. match & Realism & M-F$_1$ & Hate F$_1$\\ 
\midrule 
OPT & FT & Y & 93\% & 63\% & 66\% & .783 & .683\\
 &  & N & N/A & / & 0\% & .774 & .652\\
 &FS & Y & 90\% & 39\% & 83\% & .782 & .691\\
 &  & N & 81\% & / & 70\% & .791 & .700\\
 \midrule
OPT-IML & FT & Y & 96\% & 53\% & 66\% & .789 & .681 \\
 &  & N & N/A & / & 0\% & .796 & .690\\
 &FS & Y & 90\% & 57\% & 79\% & .789 & .698\\
 &  & N & 81\% & / & 73\% & .792 & .699\\
 \midrule
T5 & FT & Y & 83\% & 59\% & 80\% & .792 & .696\\
 &  & N & 74\% & / & 30\% & .789 & .684\\
 &FS & Y & N/A & N/A & 0\% & .786 & .682\\
 &  & N & N/A & / & 0\% & .798 & .700\\
 \midrule
Flan-T5 & FT & Y & 94\% & 66\% & 81\% & .792 & .696\\
 &  & N & 74\% & / & 41\% & .793 & .691\\
 &FS & Y & 89\% & 36\% & 84\% & .786 & .684\\
 &  & N & 87\% & / & 86\% & .774 & .662\\ \bottomrule
\end{tabular}
}
\caption{Generated texts labeled as correct by human annotators in terms of labels, target categories, and realism. N/A refers to cases in which all of the generated texts were nonsensical (0\% realistic), with impossible assignment of labels or categories. We also report the model performance from Table \ref{tab:deberta-results-gen} in terms of Macro-F$_1$ and Hate F$_1$, in order to make comparisons between model performance and manual annotation results easier.}
\label{tab:manual analysis}
\end{table}

An overview of the manual annotations is reported in Table \ref{tab:manual analysis}. 
In most cases, the addition of target information results in more realistic texts and, in general, more accurate label assignment by the generation model. However, this is not directly associated with the augmented data improving model performance when used for training. For instance, the setting that yields the best results with data generated by T5 (0.798 M-F$_1$ and 0.700 Hate-F$_1$, see Table \ref{tab:deberta-results-gen}) is the one with few-shot prompting without target information, whose generated sentences are deemed as never realistic by the human annotators. On the other hand, the worst classification setting is obtained with examples generated by OPT using finetuning and no target information (0.774 M-F$_1$ and 0.652 Hate-F$_1$), which led the model to generate nonsensical texts.

If we compare the behavior of the different generative models, we observe that Flan-T5 is the most consistent model in terms of realistic generated text, being able to produce some realistic sentences in every setting, and obtaining the highest Realism score overall. OPT and OPT-IML, on the other hand, generate nonsensical texts when using finetuning without target information, while T5 does not generate any realistic sentence when few-short prompting is used, both with and without target information. 

Overall, the rate of realistic texts and the accuracy of the identity categories are still somewhat low compared to the correctness of label assignment, showing that the generative models we tested might have difficulties dealing with more than one type of constraint/instruction. Indeed, while few-shot (FS) approaches tend to lead to more realistic generated sequences (apart from T5), this typically entails lower label correctness or target match, and vice-versa.

\begin{table*}[ht]
\centering
\resizebox{0.8\linewidth}{!}{%
\addtolength{\tabcolsep}{-0.2em}
\begin{tabular}{@{}ccc|ccccccc@{}}
\toprule
 &  &  & Women & Trans p. & Gay p. & Black p. & Disabled p. & Muslims & Immigrants \\
 \midrule
\multicolumn{3}{l|}{No Augmentation} & .142$^{.05}$ & .101$^{.03}$ & .252$^{.06}$ & .216$^{.07}$ & .113$^{.04}$ & .147$^{.04}$ & .109$^{.01}$ \\
\multicolumn{3}{l|}{EDA} & {\ul .400}$^{.04}$ & {\ul .485}$^{.09}$ & {\ul .590}$^{.06}$ & {\ul .643}$^{.09}$ & {\ul .463}$^{.11}$ & {\ul .546}$^{.13}$ & {\ul .420}$^{.06}$ \\
\midrule
Model &  & Target &  &  &  &  &  &  &  \\
\midrule
\multirow{4}{*}{\begin{tabular}[c]{@{}c@{}}OPT\\ +\\ EDA\end{tabular}} & \multirow{2}{*}{FT} & Y & \textbf{.458$^{.12}$} & \textbf{.526$^{.12}$} & \textbf{.646$^{.10}\diamond$} & \textbf{.671$^{.09}$} & \textbf{.533$^{.13}\diamond$} & \textbf{.608$^{.16}$} & \textbf{.529$^{.18}\diamond$} \\
 &  & N & .354$^{.10}$ & .394$^{.13}$ & .537$^{.14}$ & .581$^{.15}$ & .372$^{.10}$ & .538$^{.17}$ & .402$^{.13}$ \\
 & \multirow{2}{*}{FS} & Y & .384$^{.15}$ & .412$^{.15}$ & .552$^{.10}$ & .605$^{.10}$ & .408$^{.15}$ & .511$^{.19}$ & .411$^{.21}$ \\
 &  & N & .313$^{.10}$ & .316$^{.10}$ & .464$^{.16}$ & .497$^{.13}$ & .324$^{.10}$ & .456$^{.16}$ & .350$^{.14}$ \\
 \midrule
\multirow{4}{*}{\begin{tabular}[c]{@{}c@{}}OPT- IML\\ + \\ EDA\end{tabular}} & \multirow{2}{*}{FT} & Y & .409$^{.11}$ & .468$^{.18}$ & .583$^{.16}$ & .612$^{.14}$ & .493$^{.15}$ & .572$^{.19}$ & .488$^{.19}$ \\
 &  & N & .337$^{.09}$ & .369$^{.14}$ & .517$^{.16}$ & .531$^{.19}$ & .355$^{.14}$ & .525$^{.20}$ & .370$^{.16}$ \\
 & \multirow{2}{*}{FS} & Y & .396$^{.14}$ & .415$^{.12}$ & .565$^{.06}$ & .632$^{.06}$ & .403$^{.14}$ & .545$^{.15}$ & .452$^{.18}$ \\
 &  & N & .324$^{.05}$ & .315$^{.06}$ & .436$^{.12}$ & .527$^{.12}$ & .321$^{.11}$ & .415$^{.14}$ & .308$^{.09}$ \\
 \midrule
\multirow{4}{*}{\begin{tabular}[c]{@{}c@{}}T5\\ +\\ EDA\end{tabular}} & \multirow{2}{*}{FT} & Y & .305$^{.05}$ & .299$^{.12}$ & .470$^{.13}$ & .472$^{.11}$ & .323$^{.11}$ & .412$^{.06}$ & .318$^{.08}$ \\
 &  & N & .273$^{.07}$ & .273$^{.07}$ & .502$^{.08}$ & .518$^{.10}$ & .309$^{.06}$ & .417$^{.12}$ & .303$^{.08}$ \\
 & \multirow{2}{*}{FS} & Y & .357$^{.08}$ & .382$^{.13}$ & .518$^{.16}$ & .547$^{.16}$ & .341$^{.11}$ & .527$^{.18}$ & .388$^{.15}$ \\
 &  & N & .402$^{.13}$ & .457$^{.16}$ & .594$^{.14}$ & .620$^{.14}$ & .436$^{.18}$ & .580$^{.18}$ & .478$^{.18}$ \\
 \midrule
\multirow{4}{*}{\begin{tabular}[c]{@{}c@{}}Flan-T5\\ +\\ EDA\end{tabular}} & \multirow{2}{*}{FT} & Y & .287$^{.06}$ & .257$^{.08}$ & .447$^{.12}$ & .454$^{.10}$ & .254$^{.08}$ & .436$^{.11}$ & .294$^{.09}$ \\
 &  & N & .300$^{.05}$ & .301$^{.08}$ & .449$^{.13}$ & .456$^{.12}$ & .307$^{.09}$ & .475$^{.16}$ & .337$^{.11}$ \\
 & \multirow{2}{*}{FS} & Y & .371$^{.13}$ & .428$^{.09}$ & .576$^{.10}$ & .613$^{.11}$ & .404$^{.10}$ & .567$^{.15}$ & .441$^{.13}$ \\
 &  & N & .388$^{.13}$ & .411$^{.16}$ & .556$^{.14}$ & .593$^{.15}$ & .399$^{.16}$ & .543$^{.20}$ & .422$^{.17}$ \\
 \bottomrule
\end{tabular}
}
\caption{DeBERTa results on HateCheck (hate-F$_1$) by target identity, averaged across 5 runs. \textit{p.} is an abbreviation for \textit{people}. Statistical significance is calculated against the results obtained with EDA. $\diamond$: statistically significant ($\tau=0.5$).}
\label{tab:results-deberta-hatecheck}
\end{table*}

\subsection{HateCheck Analysis}
\label{sec:hatecheck}
We perform a second qualitative analysis using the HateCheck test suite \citep{rottger-etal-2021-hatecheck}, a collection of functional testing examples that enable targeted diagnostic insights of hate speech detection models. We focus on the models trained with augmented data using generative DA + EDA for this analysis (Table \ref{tab:deberta-results-gen+eda}), since they yield the best classification performance. Again, each generative model + EDA is used in four settings to generate new data: with finetuning or few-shot prompting, each one with or without target information. 

All HateCheck test cases mention a specific target identity, to allow the exploration of unintended biases against different target groups. However, the target groups used in HateCheck do not fully overlap with the target identity groups in the MHS corpus (Figure  \ref{fig:distr-mhs}). 
The target identities that are present in HateCheck are: women (which would fall under \textit{gender} in MHS), trans people (\textit{gender} in MHS),
gay people (\textit{sexuality} in MHS), black people (\textit{race} in MHS),
disabled people (\textit{disability} in MHS), Muslims (\textit{religion} in MHS) and
immigrants (\textit{origin} in MHS). The \textit{age} category is  present in MHS corpus and entirely missing in HateCheck.

We test the models trained on generation and EDA-augmented data (see setup in Section \ref{sec:mixture}) on HateCheck targets, reporting the performance on a per-target basis in terms of hateful-class F$_1$. The results are reported in Table \ref{tab:results-deberta-hatecheck}, divided by the target identity categories used in HateCheck. We also report, as baselines, the results of the classification model using no augmentation and EDA alone. 

Overall, we notice significant improvements over the \textit{no augmentation} baseline across setups and target identities, although these improvements are variable. Indeed, EDA + OPT finetuned (FT) with target information lead to performance increases of up to .109 hate-F$_1$ over EDA alone, also resulting in more evenly distributed scores across targets, potentially showing less fairness-related weaknesses, paired with an improvement in performance across \textit{all targets}. Interestingly, however, even overlapping targets between the MHS dataset and HateCheck show different improvements across generative DA setups. For instance, the HateCheck target \textit{disabled people} apparently benefits from target information inclusion more than the \textit{disability} category in the MHS corpus, suggesting that the inclusion of identity group information could affect performances differently in and out of domain. 

Furthermore, HateCheck test instances are manually created starting from templates that include specific mentions of identity-group terms, which could lead to sentences that are very different from `real' examples. For instance, one of the hateful templates used in HateCheck is `\textit{I hate [IDENTITY]}', in which \textit{[IDENTITY]} is substituted with \textit{women}, \textit{disabled people}, and so on for all of the targets reported above. This can result in somewhat stereotypical examples, while hateful posts in the MHS corpus often contain slurs and offensive terms instead of  neutral names to refer to people belonging to a certain group.

From the distribution of HateCheck scores, it appears that the OPT models tend to benefit from both finetuning and the addition of target information, while T5 and Flan-T5 tend to lead to the best gains when prompted few-shot, with the influence of target information insertion having different effects on distinct target identity groups. The effect of finetuning, prompting, and target information therefore seems to depend more on the type or architecture of the generative model rather than whether it is instruction-tuned or not, as OPT-IML and Flan-T5, which are both instruction-tuned, exhibit different trends.

Looking specifically at the HateCheck counterparts of the targets that are least represented in the MHS corpus (\textit{disabled people} for \textit{disability} and \textit{Muslims} for \textit{religion}), it is clear that data augmentation with target information tends to lead to consistently better results than the non-target-aware data augmentation approach with OPT and OPT-IML, while for both T5 models this is not always the case. This might indicate that, as previously discussed, models might struggle with multiple constraints when generating new examples.

Finally, in terms of fairness, it appears that data augmentation using generative models and EDA can work towards improving the performance of models on all targets included in HateCheck, even if there is no 1:1 mapping with the original targets. This indicates that this approach can potentially be effective in improving the performance of models across different targets of hate. 

\section{Conclusions}\label{conclusion}
We have investigated the impact of data augmentation with generative models on specific targets of hate, experimenting with instruction-finetuned models and the addition of target information when generating new sequences. Our main goal is to analyze which approaches work best to cover targets of hate speech that have been neglected in research on the topic. With our approach we therefore aim at helping the development of fairer systems, that give equal recognition to the different minority groups usually targeted by online haters. 

Overall, it appears that DA methods
have different types of impact on different targets, but they can improve performance even for scarcely represented identity categories \textbf{(Q1)}. 
We observed that generative data augmentation alone is not as strong as simpler methods such as EDA, both globally and on a per-target basis, especially given that generative DA is highly computationally expensive. However, their combination can lead to models that are more robust, especially for more scarcely represented target identities, highlighting the potential of this type of approach \textbf{(Q3)}.
Through a qualitative analysis, we also emphasized the fact that including target information when generating synthetic examples can facilitate the creation of examples that are more realistic and exhibit more correct label assignments \textbf{(Q2)}, although further work could investigate why these characteristics do not directly correlate with downstream task performance.

\section*{Limitations}
\label{limit}
The current work inevitably has some limitations that may affect its impact. First of all, we cast target groups as non-overlapping categories, while we could also consider intersectionality by generating messages targeting two or more identity groups.

In addition, we worked on English data because of the availability of the Measuring Hate Speech corpus, which is large enough to perform our DA experiments and presents the kind of fine-grained target annotation required in our study. However, we are aware that DA would benefit more classification with lower-resourced languages. Furthermore, in order to be able to generalize our findings, we should ideally test them also on other datasets. 

One potential downside of generation-based DA is that it can increase the over-reliance of models on identity terms \citep{casula-tonelli-2023-generation}, so biases could be propagated in the augmentation process. However, by generating target-specific examples also for the non-offensive class, we ideally aim at implicitly contrasting identity term bias.
We acknowledge that the same techniques for data augmentation, able of creating thousands of hateful examples to attack different target groups, could also be used with the goal of hurting the same groups that we want to support.\footnote{Because of this, we provide all the necessary details for the reproduction of our results, but we do not plan to openly release the code or to upload the generated data produced by our experiments, especially in order to avoid it being crawled and ending up in the training data of LLMs in the future. We are, however, open to sharing the data and code with other researchers who might be interested.}
Finally, our approach requires at least some manual annotation to create the initial set of training data. We would like to investigate in the future the possibility to adopt DA for zero-shot approaches, especially given that instruction tuning has been found effective especially in zero-shot scenarios.

\section*{Impact statement}
Our findings have several potential implications. From a research point of view, we  show how differently hate speech classifiers perform when dealing with different targets of hate. This suggests that, beside developing approaches to hate speech detection that are robust across languages, domains and platforms \citep{pamungkas2023towards}, the specific targets of hate and their representation in datasets should be another dimension to investigate, because it can greatly affect classification performance. We advocate not only for the creation of datasets that cover equally a diverse set of hate targets, but also for the development of target-specific ones, including the categories that have been previously neglected and would benefit from more data-driven insights such as religious minorities and disabled people \citep{Silva_Mondal_Correa_Benevenuto_Weber_2021}.

Concerning the use of generative language models for data augmentation, we show that it can be beneficial, even when using only openly available models. However, given their high computational costs, also alternatives like EDA could be considered if limited resources are available, because they can still yield performance improvements compared to a low-resource setting. Again, no one-fits-all solution or approach to generation should be pursued. In general, while we acknowledge that generative language models have contributed to reaching state-of-the-art performance in several NLP tasks, we also argue that more thorough comparisons and evaluations are needed, including qualitative ones, especially when dealing with sensitive topics like online hate speech. For instance, it would be interesting to further investigate why synthetic texts that are manually labeled as realistic do not necessarily improve classification performance, and also to assess whether it applies to other tasks.

Ultimately, the goal of our work is to contribute to the development of AI systems for online content moderation that are fairer and more inclusive \citep{XU2024103665}. Several works have identified biases, especially in large language models, and have proposed approaches to mitigate them \citep{bender2021-stochasticparrots,SONG2023103277}. Our contribution shares the same overall goal, but focuses on the representativeness of diverse hate targets, aiming at developing systems that can detect equally well offenses and hate speech against different communities. 

\bibliography{anthology,custom}

\end{document}